\documentclass[12pt,english]{article}

\usepackage{amssymb}
\usepackage{polski}
\usepackage{ifthen}
\selecthyphenation{polish}
\mathchardef\ogon="012C%
\newcommand{\as}{a\kern-0.22em\lower.40ex\hbox{$_{\ogon}$}}
\newcommand{\As}{A\kern-0.22em\lower.40ex\hbox{$_{\ogon}$}}
\newcommand{\es}{e\kern-0.24em\lower.40ex\hbox{$_{\ogon}$}}
\newcommand{\Es}{E\kern-0.22em\lower.40ex\hbox{$_{\ogon}$}}
\makeatletter

\usepackage[a4paper, left=2.5cm, right=2.5cm, top=2.5cm, bottom=3.5cm, headsep=1.2cm]{geometry}
\newtheorem{theorem}{Theorem}[section]

\newtheorem{algorithm}[theorem]{Algorithm}

\newtheorem{definition}[theorem]{Definition}

\newtheorem{remark}[theorem]{Remark}

\def\qed{\hbox to 0pt{}\hfill$\rlap{$\sqcap$}\sqcup$}
\makeatother
\usepackage{amssymb}
\usepackage{amsmath}
\usepackage{amsfonts}
\usepackage{color}
\usepackage{graphicx}
\usepackage{graphics}
\numberwithin{equation}{section}
\usepackage{babel}
\date{}
\title{A comprehensive study of clustering a class of 2D shapes}

\author{Agnieszka Kaliszewska and
Monika Syga }

\usepackage{graphicx}
 \usepackage{mathptmx}      
\usepackage{latexsym}
\usepackage[T1]{fontenc}
\usepackage{latexsym}
\usepackage{amssymb}
\usepackage{amsmath}
\usepackage{color}
\usepackage{xcolor}
\usepackage{graphicx}
\usepackage{graphics}

\begin{document}
	\maketitle

\begin{abstract}
 The paper concerns clustering   with respect to the  shape and size  of 2D contours that are boundaries of cross-sections of 3D objects of revolution. We propose a number of similarity measures based on combined disparate Procrustes analysis (PA) and  Dynamic Time Warping (DTW) distances. Motivation and the main application for this study comes from archaeology.  The performed computational experiments refer to the  clustering  of archaeological pottery.
 \end{abstract}
 
 
\section{Introduction}
\label{intro}

We investigate clustering  of 2D cross-sections of
3D objects of revolution with respect to  shape and size.
We do not tackle the problem of recognizing whether a given 3D object is rotationally symmetric or not.

Clustering  is used extensively in unsupervised machine learning applications such as information retrieval and natural language  understanding as well as in numerous fields such as economics, biology, medicine, physics, e.g. \cite{AMCS1}. However, it evades an in-depth unified framework. This is mostly due to specific challenges arising depending on the nature of data, resulting in case-specific algorithms. Some of these challenges have been presented in e.g. \cite{kleinberg} and \cite{Nino_Berzal}, There, the so called {\em impossibility theorem} is proved which states that no clustering algorithm exists satisfying the following three axioms: scale invariance, richness, and consistency. For further discussion see e.g.,   \cite{Addad_Kanade}, \cite{Cohen_Addad}, \cite{Davidson}.

 To obtain significant results for clustering of 2D cross-sections of objects of revolution with respect to shape, and size, aside from the choice of the algorithm, the following issues are of fundamental importance: the shape concept, the data representation, and the  similarity measure \cite{Wierzchon_Klopotek} and \cite{Jain}. In our investigation we address those three issues in the context of clustering of archaeological pottery.
 
\subsection{State-of-the-art}
The problem of automatic and semi-automatic classification of archaeological pottery has been investigated by several authors. In \cite{Piccoli_2015} similarity measure is proposed based on the extraction of shape features from the silhouette through medialness. In \cite{sablatnig_1998} the authors make use of the 3D models of the investigated shapes in combination with syntactic pattern recognition approach. The classification of 3D shapes, and the issue of matching a fragment to its fully preserved counterpart, is investigated in \cite{maiza_2005}, where the shape information is extracted based on a skeleton. In \cite{hristov_agre} a  classification, based on contour representations and the representative functions, is presented. Most recently, a comprehensive overview of classification and shape matching methods in the study of archaeological ceramics has been published in \cite{WILCZEK}.
Procrustes analysis and generalized Procrustes Analysis appear in the analysis of archaeological material as stand-alone methods, for example in \cite{dryden}, there they are used to study the variability of landmark points in the study of shapes of a class of objects.

On the other hand, clustering methods in archaeological applications are scarce. One such method for shape contours clustering that uses the representative functions for shape representation and the Euclidean distance to define the similarity measure is presented in \cite{gilboa_karasik_2004}. Other contributions on the clustering of archaeological pottery include 
\cite{Mumford_2002,Mumford1991,Mumford_2D} and the references therein. 
In the paper, \cite{C&C} we have proposed a similarity measure based on Dynamic Time Warping (DTW) to compare representative functions as defined in \cite{gilboa_karasik_2004}. The performed experiments showed that DTW could provide a promising new similarity measure. This observation formed a basis for introducing the similarity measures as defined in the present paper.

In the present investigation, we use the concept of shape introduced by \cite{kendall_1977}, \cite{kendall1989}, Procrustes Analysis (PA) and DTW to propose new weighted similarity measures. Up to our knowledge, this approach has not yet been used in the context of clustering of archaeological pottery. It is worth mentioning that our approach can be applied to clustering of any 3D objects of revolution for which the revolution axis and the 2D section are known.

\subsection{The methods}
The concept of shape we adopt, and a formal basis of shape analysis in the form of the so-called Kendall shape goes back to a series of publications by David Kendall \cite{kendall_1977}, \cite{kendall1989}, \cite{Goodall}.
 Kendall shape analysis in turn is inspired by the shape theory introduced by Karol Borsuk in the sixties and seventieth \cite{borsuk_dydak_1980} and is motivated by applications in archaeology. For the systematic exposition of the topic see the monograph \cite{da2010shape}.

Informally, the shape of an object $X$ is defined as all the geometrical information that remains when location, scale and rotational effects are filtered out from the object $X$, or, in other words, the shape is the geometry of an object $X$ modulo
position, orientation, and size. In consequence,  the shape space is the space of equivalence classes defined by a given class of objects $X$.

 Such understanding of shape gives rise to the concept of the {\em Procrustes distance} or {\em Procrustes distance measure} (see e.g. \cite{Goodall})  between the objects $X_{1}$ and $X_{2}$ as the distance of $X_{1}$ to the equivalence class defined by $X_{2}$. The Procrustes distance is not a distance in the formal meaning of the term. 
Based on Procrustes distance, Procrustes-type similarity measures between shapes are proposed. 

In numerous applications, PA and  Procrustes-type similarity measures are  combined with other similarity measures,  e.g. in \cite{8545040}, combining PA with the Principal Component Analysis (PCA) is proposed to investigate  3D gait recognition problem, \cite{eguizabal2019procrustes}, 
PA is combined with the DTW to investigate surgery task classification in \cite{Albasri}.

DTW is devised to compare time series. More about this topic can be found \cite{Aronov2006} and \cite{Efrat_2007}.  In general, similarity measures devised with the help of   DTW can be applied to any objects composed with linearly parametrized (ordered) elements (components).

In our investigations, we define similarity measures by combining and Procrustes based similarity measures with DTW similarity measures. The computational experiments confirm the efficiency of the proposed approach.

2D cross-sections are represented through their contours, i.e 2D curves that are boundaries of these 2D cross-sections. In archaeological applications, those 2D curves are generated from technical drawings. This means that their position on the plane is not random and is determined by a set of strict rules. As a consequence, the problem of possible misalignment is not as crucial as in other applications, e.g. \cite{missaligned_2010,joint_clustering_2012}. By applying the boundary-based representation of cross-sections, the problem of clustering of 3D objects of revolution is reduced to the clustering of 2D curves with a given revolution axis with respect to shape and size. The problem of shape and size analysis of 2D curves and functions is gaining increasing interest due to many important applications, such as cardiovascular analysis \cite{missaligned_2010}, finance interest rates \cite{finance_curves}, nuclear industry \cite{nuclear_curves}.

A specific feature of clustering of archaeological pottery is that the contours should be classified with respect to subtle differences in shapes and size. Hence, the adopted similarity measures and the resulting clusters should reflect as precisely as possible, those subtle differences. The subtle differences are of great importance in the clustering of contours, as a relatively small change in the contour might have a great impact on the shape of the respective 3D object of revolution.

\subsection{The aim and the contribution}
\label{aim}

The main aim of the present investigation is to propose weighted similarity measures, based on PA  and DTW, for clustering of 2D curves defining boundaries of  2D cross-sections of 3D  objects of revolution, with respect to shape and size. 

Depending upon the particular case, the contribution of shape and size in the overall similarity measure is adjusted by proper choice of weights.  

The secondary aim is to apply these similarity measures to the automatic generation of typologies (clusters) for archaeological pottery fragments. Clustering archaeological pottery is considered a process that may depend, to some extent, upon subjective judgements. This subjective judgement can be reflected in the choice of weights in the similarity measure.

The contribution  of the paper is as follows.
\begin{enumerate} 
	\item By using the PA (formula \eqref{procrustes_measure}) and DTW (formula \eqref{dtw_3})  we propose   two novel  curve-based pair-wise formulas to compare two curves:  
		\begin{description}
		\item[-] Direct Composition similarity measure DC (see formula \eqref{directcomposition}),
		\item[-] scale (size) factor $\gamma^*$ (formula  \eqref{procrustes_measure}).
		\end{description}
		
	\item We use direct composition similarity measure DC (formula \eqref{directcomposition}), scale similarity measure $\gamma^*$ (formula \eqref{pa}) and the procrustes similarity measure PA (formula \eqref{procrustes_measure}) to generate various similarity matrices with the help of weights and normalization. By  introducing weights and  normalizations we get the following similarity matrices for which we perform our experiments:
		\begin{enumerate}
		    \item PSM (formula \eqref{psm})-Procrustes similarity matrix, emphasizes the variations in the shape of  curves in data set,
		    \item DCM (formula \eqref{dcm})-direct composition matrix, emphasizes the variations in shape and size of  curves in data set,
		    \item SCM (formula \eqref{scm})-scale component matrix, emphasizes only the variations in the size of  curves in data set,
		    \item WPSM (formula \eqref{wpsm})-weighted Procrustes and scale component matrix, emphasizes the variations in shapes and size of  curves in data set, by changing the weights we can decide which feature we want to highlight in a given experiment,
		    \item WNDCSM (formula \eqref{wndc})-weighted direct composition and scale component matrix where direct composition values are normalized. Since the direct composition values are usually much higher than the values of scale component we use normalization.
		    \end{enumerate}
		\item We apply the proposed similarity measures to the generation of typologies of archaeological pottery. The computational experiment is based on several data sets, some of them real-life data, larger data sets are generated automatically by augmentation procedures.
	\end{enumerate}

\subsection{The organization of the paper}
The organization of the paper is as follows. In Section \ref{section_classofobjects} we describe the class of  investigated objects and 
we discuss the  representations of 2D shapes (contours). In Section \ref{section_similarity_measures} we present basic elements of PA and DTW and we define  direct composition similarity measure DC. In Section \ref{section_proposedscheme} we  introduce similarity measures (a)-(e) and present  a general algorithm. In Section \ref{section_application} we describe the archaeological objects on which we test our method. In Section \ref{section_experiment} we describe the numerical experiments conducted and the data used.  In Section \ref{results_discussion} we discuss the obtained results.  Section \ref{section_conclusions}  concludes the paper.

\section{The class of investigated objects}
\label{section_classofobjects}

Geometrically, 3D  objects of revolution is a solid obtained by rotating a plane curve around a straight line (axis of revolution) that lies on the same plane.
We investigate  3D objects of revolution obtained by rotating their 2D cross-sections,  (not necessarily curves) around an axis of revolution. Boundaries of 2D sections are called contours.
\begin{figure}[h]
	\centering
	\includegraphics[width=0.5\textheight]{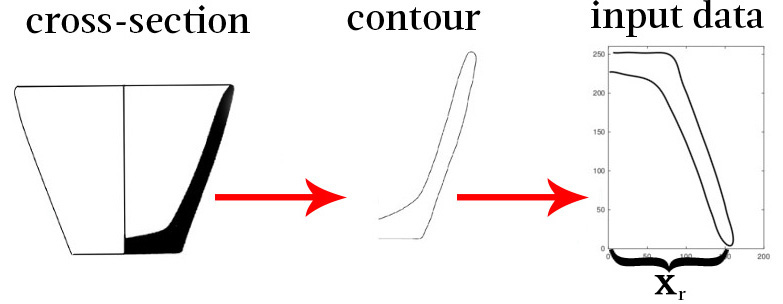}
	\caption{The transformation of a section of rotationally symetric object into a 2D input data curve.}
	\label{fig:1}
\end{figure}
We assume each contour is a 2D open curve $\alpha\subset\mathbb{R}^{2}$. 
  Each curve 
is a function $\alpha: [t_{0},t_{1}] \rightarrow\mathbb{R}^{2}$ represented as 
\begin{equation}
\alpha(t):=(x(t),y(t))\ \ \ t\in[t_{0},t_{1}],
\label{curve}
\end{equation}
where $x:[t_{0},t_{1}] \rightarrow\mathbb{R}$, and $y:[t_{0},t_{1}] \rightarrow\mathbb{R}$.

 We assume that 
\begin{description}
\item[(A)]  the curves are contours of cross-sections of objects of revolution with the revolution axis $OY$ and are all  located in the first quadrant of the plane ($x \geq 0$, $y \geq 0$)
\item[(B)] the lowermost point (points), i.e., points  with the smallest value of the coordinate $y$ of a given curve is forced to lay on OX axis (i.e. its $y$ coordinate equals zero), and the point 
$(x_{r}, 0)$, where  $x_{r}$  is the smallest value of $x$-coordinate corresponding  to $y=0$ 
defines  the {\em the radius $r$ of the 3D object of revolution} (see Figure \ref{fig:1}), 
\begin{equation}
\label{radius}
r:=x_{r},
\end{equation} 
\item[(C)] the uppermost point (points), i.e. points with largest values of the coordinate $y$ are assumed to lie on the revolution axis $OY$.
\end{description}

\begin{remark}{}
Let us note that that assumption (A) is not limiting. In case we classify objects with various revolution axes, we can always rotate (and shift) the object so as to make the assumption (A) satisfied. Assumptions (B) and (C)  refer to the situation which is natural in the archaeological application, i.e. the locations of contours on the plane are not random and defined as in (B) and (C). In the case where the locations of contours are random, some position normalization is needed, e.g. one can apply position normalization with which   Procrustes Analysis typically starts.
\end{remark}

\subsection{Shape representations}
\label{section_similarity_measures}
 From the digitized cross-section the boundary discrete curve (contour) is extracted via standard techniques of contour extraction.

Contours are smoothed by the Savitzky-Golay filtering to  preserve as much as possible the original shape of a profile.

In further analysis,  we consider discrete curves,
i.e. a  curve  $\alpha$ is a pair of vectors,

\begin{equation} 
\label{discrete representation}
\alpha:=(x(i),y(i)),\ i=1,...,k_{\alpha}.
\end{equation}

 Consequently, in the sequel, all data are vectors of different dimensions. When the original data is acquired in the form of bitmaps (e.g. scanned hand-made drawings), the accuracy of discretization depends on the resolution of the scanned image.

\section{Similarity measures}
\label{section_similarity_measures}

We start by  recalling  PA and DTW similarity measure. 
For more detailed readings, see e.g. 
\cite{DTW_book,Efrat_2007,optimization_for_procrustes}.

 Next, in subsection \ref{section_DC}, we define new  similarity measures based on PA and  DTW, which is called direct composition DC.

\subsection{Dynamic Time Warping (DTW)}

\label{sDTW}

In view of the adopted contour representations, the compared curves $X$ and $Y$  are represented by
sequences $X :=
(x_1, x_2, . . . , x_N)$ of length $N \in \mathbb{N}$ and $Y := (y_1, y_2, . . . , y_M)$ of length $M \in \mathbb{N}$, respectively, i.e.,  $X \in \mathbb{R}^N$,  $Y \in \mathbb{R}^M$.

\begin{figure}
  \centering
  \includegraphics[width=8cm]{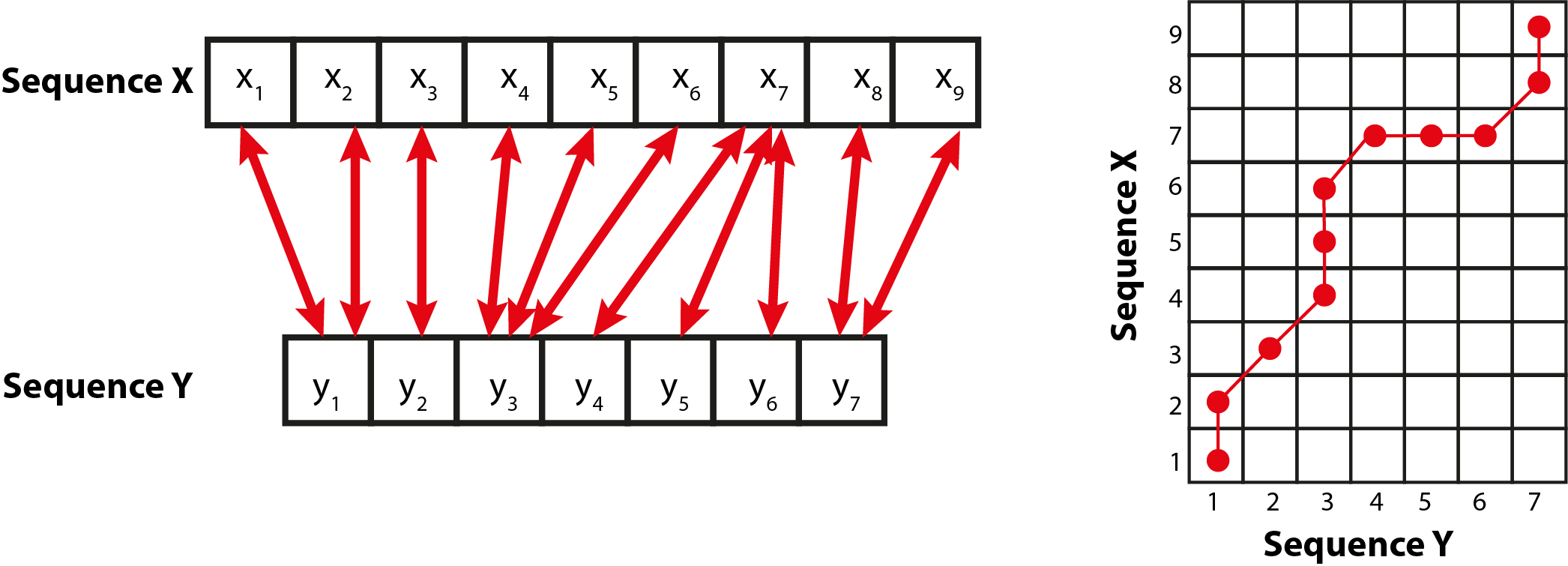}
  \caption{An example of an alignment and a warping path $p$. 
  }
  \label{fig_DTW}
\end{figure}

 To make the comparison between these sentences the DTW algorithm align them by applying the following procedure. First, the definition of a warping path is introduced.  
\begin{definition}{Definition 4.1, \cite{DTW_book}}
\label{def1}
An $(N,M)$-warping path (or simply referred to as warping
	path if $N$ and $M$ are clear from the context) is a sequence $p = (p_1, . . . , p_L)$
	with $p_{\ell} = (n_{\ell},m_{\ell}) \in \{1,...,N\} \times \{1,...,M\}$ for  $\ell\in \{1,...,L\}$
	satisfying the following three conditions.
	\begin{description}
		\item [(i)] Boundary condition: $p_1 = (1, 1)$ and $p_L = (N,M)$.
		\item [(ii)] Monotonicity condition: $n_1 \le n_2 \le  . . . \le n_L$
		and $m_1 \le m_2 \le . . . \le m_L.$
		\item [(iii)] Step size condition: $p_{\ell+1}-p_{\ell} \in \{(1, 0), (0, 1), (1, 1)\}$
		for $\ell \in \{1,...,L-1\}.$
	\end{description}
\end{definition}
Hence, an $(N,M)$-warping path $p = (p_1, . . . , p_L)$ is defined by an alignment between two
sequences $X = (x_1, x_2, . . . , x_N)$ and $Y = (y_1, y_2, . . . , y_M)$ by assigning the
element $x_{n_l}$ of X to the element $y_{m_l}$, $l\in \{1,...,L\}$ of Y. The Figure \eqref{fig_DTW} shows how an alignment is transformed into a warping path.

Now we define a cost measure which allows finding the best warping path. In the following, by $F$ we denote  the 
feature space, i.e., $x_n, y_m \in F$ for $n \in \{1,...,N\}$, $m \in \{1,...,M\}$.
To compare two different elements $x, y \in F$, a local cost measure is defined to be a
function $
c : F \times F \rightarrow \mathbb{R}_{+}$.
Evaluating the local cost measure for each pair of elements of the sequences X and Y, we obtain the cost matrix
$C \in \mathbb{R}^{N\times M}$ defined by $C(n,m) := c(x_n, y_m)$.


The total cost $c_p(X, Y )$ of a warping path $p$ between X and Y with respect
to the local cost measure c is defined as
\begin{equation}
\label{dtw_2}
c_p(X, Y) := \sum_{\ell=1}^{L} c(x_{n_l}, y_{m_l}),
\end{equation}
where, as in Definition \ref{def1},  $p = (p_1, . . . , p_L)$  with $p_{\ell}=(x_{n_{\ell}},y_{m_{\ell}})$, $\ell\in \{1,...,L\}$.
An optimal warping path between X and Y is a warping path
$p^*$ having minimal total cost from among all possible warping paths. The DTW
similarity measure $DTW(X, Y )$ between X and Y is then defined as the total cost of $p^*$:
\begin{equation}
\label{dtw_3}\begin{array}{l}
DTW(X, Y ) := c_{p^{*}} (X, Y ) \\
= \min\{c_{p}(X, Y ) \ |\  p \text{ is an $(N,M)$-warping path}\}.\end{array}
\end{equation}

The DTW similarity measure is akin to Fr\'echet distance, see \cite{Aronov2006,DTW_book}. In our experiments we use the DTW function from Matlab Signal Processing Toolbox. 

\subsection{Procrustes analysis (PA)}
\label{section_PA}

Procrustes Analysis (PA) allows to perform the statistical shape analysis and  to compute a similarity measure for any given pair of 2D curves represented  in the  discretised form \eqref{discrete representation} see e.g. \cite{procrustes_problems,optimization_for_procrustes}.

Let us consider the set of discrete curves  $\alpha:=(x(j),y(j))_{j=1}^{k}\in(\mathbb{R}\times\mathbb{R})^{k}$ (regarded as row vectors). Let ${\cal T}$ be the set of similarity transformations defined on $(\mathbb{R}\times\mathbb{R})^{k}$.  ${\cal T}$ is given in the form of triples $T:=(\gamma, R, t)$, where:
\begin{description}
\item[-] $\gamma\in\mathbb{R}_{++}$ represents a (uniform) scaling factor and defines the scaling transformation $\gamma:(\mathbb{R}\times\mathbb{R})^{k} \rightarrow (\mathbb{R}\times\mathbb{R})^{k}$, $\gamma(\alpha)= [\gamma (x(1),y(1)),..., \gamma(x(k), y(k))]$,

\item[-]$R$ is a $2\times 2$ rotation matrix of the angle $\theta$, and defines rotation transformation $R:(\mathbb{R}\times\mathbb{R})^{k} \rightarrow (\mathbb{R}\times\mathbb{R})^{k}$,
$$
R=\left[\begin{array}{cc}
\cos\theta&-\sin\theta\\ \sin\theta&\cos\theta\end{array}\right],
$$
$R(\alpha)=R\alpha$,
\item[-]$t\in (\mathbb{R}\times\mathbb{R})^{k}$ is a translation vector, which defines the transformation $t:(\mathbb{R}\times\mathbb{R})^{k}\rightarrow (\mathbb{R}\times\mathbb{R})^{k}$, $t(\alpha)=t+\alpha$, with components $t(j)\in \mathbb{R}\times\mathbb{R}$, $j=1,...,k$.
\end{description}


For the input discrete curves 
$\alpha_{1}:=(x_{1}(j),y_{1}(j))=(z_{1}(j))$, $j=1,...,k$, $\alpha_{2}:=(x_{2}(j),y_{2}(j))=(z_{2}(j))$,  $i=1,...,k$, 
the Procrustes similarity measure $PA(\alpha_{1},\alpha_{2})$ is defined as 


\begin{equation} 
\label{procrustes_measure}
\begin{array}{l} 
PA(\alpha_{1},\alpha_{2}):=\\
\inf_{T\in{\cal T}}\ \sum_{j=1}^{k}\|z_{1}(j)-(
\gamma (R \alpha_{2})(j)+t(j))\|_{2}^{2},
\end{array}
\end{equation}
where $\|\cdot\|_{2}$ is the Euclidean distance in $\mathbb{R}^{2}$.
The procrustes similarity measure is not a distance since in general 
$PA(\alpha_{1},\alpha_{2})\neq PA(\alpha_{2},\alpha_{1})$.

Formula \eqref{procrustes_measure} can  be interpreted in terms of equivalence classes $[\cdot]$ of some equivalence relation. Namely, a (discrete) curve $\alpha:=(x(j),y(j))_{j=1}^{k}\in(\mathbb{R}\times\mathbb{R})^{k}$ belongs to the equivalence class defined by $\alpha_{1}=(x_{1}(j),y_{1}(j))_{j=1}^{k}\in(\mathbb{R}\times\mathbb{R})^{k}$, $\alpha\in[\alpha_{1}]$ iff $(x(j),y(j))=\gamma\cdot R(x_{1}(j),y_{1}(j))+t(j)$, $t(j)\in \mathbb{R}\times\mathbb{R}$, $j=1,...,k$, where $\gamma>0$ is a scaling factor, $R$ is a $2\times 2$ rotation matrix of the angle $\theta$, as defined above,
and $t\in \mathbb{R}^{2}$. In this way an equivalence relation $\equiv$ is defined as
\begin{equation}
\label{equivalence}
\alpha_{1}\equiv\alpha_{2} \Leftrightarrow (x_{1}(j),y_{1}(j))=\gamma\cdot R(x_{1}(j),y_{1}(j))+t(j),
\end{equation}
$j=1,...,k$ for some scaling factor $\gamma>0$, rotation matrix $R$ and translation vector $t$ with components $t(j)\in \mathbb{R}\times\mathbb{R}$, $j=1,...,k$. 

In our experiments we use the Procrustes function from Matlab Statistics and Machine Learning Toolbox.


\subsection{Direct Composition}
\label{section_DC}

 In the present subsection we introduce a new similarity measure which is based on composition of PA and DTW.

Let two (discrete) curves $\alpha_1,\alpha_2\in (\mathbb{R}\times\mathbb{R})^{k}$ be given. The objective function of the optimization problem \eqref{procrustes_measure} is strictly convex and bounded from below. 
By solving problem \eqref{procrustes_measure}, we obtain the minimal value $PA(\alpha_1,\alpha_2)=d$  and  the curve $Z=T^{*}\alpha_2$, where $ T^{*}\in{\cal T}$ is an optimal solution to problem \eqref{procrustes_measure}  i.e.
\begin{equation}
\label{pa}
\begin{array}{l}
PA(\alpha_1,\alpha_2)=d=\inf\limits_{T\in{\cal T}}\ \|\alpha_1-T\alpha_2\|^2\\=\|\alpha_1-Z\|^2
=\|\alpha_1-T^{*}\alpha_2\|^2,
\end{array}
\end{equation}
and  $Z:=T^{*}\alpha_2=\gamma^{*}R^{*}\alpha_2+t^{*}$, where $\gamma^{*}>0$ is the (optimal) scaling  factor and $R^{*}$ is the (optimal) rotation matrix, and $t^{*}$ is the (optimal) translation vector. 
In our experiments the curve $Z$ is taken from the Procrustes function in Matlab  Statistics and Machine Learning Toolbox.

With the help of the curve $Z$ we define the new similarity measure, called
 {\em direct composition DC} and given as
\begin{equation} 
\label{directcomposition}
DC(\alpha_1,\alpha_2)=DTW(\alpha_1,Z).
\end{equation}

Direct composition DC combines global shape similarity measure PA with DTW which takes into account local shape variability as well.

The motivation for the combination of PA and DTW measures stems from previous works, e.g. \cite{C&C} that have brought to light a necessity to introduce a similarity measure that would take into account the subtle nature of the differences between the investigated shapes. Not without importance is the possibility of introducing weights that account for the expert's knowledge, and bring his expertise into the process. 

We consider PA and DTW methods complementary, ass they address different characteristics of the shape. PA is, in our case, a robust tool for the alignment of shapes and the detection of overall shape similarity. It performs very well when used to detect scale and orientation invariant similarity between significantly different shapes. DTW on the other hand is well suited to detect subtle changes in objects that could be generally defined as similar. It fails, however, to quantify the similarity when the shapes differ in attributes such as rotation, or scale.


The idea of combining PA and DTW  in similarity measures 
related to the  analysis of temporal alignment of human motion already appeared  e.g. in \cite{Feng}.

\section{Clustering of curves - proposed algorithm}
\label{section_proposedscheme}

In the present section we propose an  algorithm for clustering of 2D contours of cross-sections of 3D objects of revolution. We assume that the given set of contours  consists of $n\geq 2$ elements. Let  $i=1,...,n$ and $j=1,...,n$, by $i$ and $j$ we denote the $i$-th contour and $j$-th contour respectively. First we calculate the ,,distances'' between two contours $i$ and $j$, $i\neq j$   by using the following three formulas
\begin{equation}
\label{padist}
    pa(i,j)=\max\{PA(i,j), PA(j,i)\},
\end{equation}
\begin{equation}
\label{dcdist}
dc(i,j)=\max\{DC(i,j), DC(j,i)\},
\end{equation}
\begin{equation}
\label{scaldist}
\gamma(i,j)=1-\min\{\gamma^*(i,j), \gamma^*(j,i)\},
\end{equation}
where $\gamma^*(i,j)$ is the optimal scaling factor obtained from the $PA(\alpha_i, \alpha_j)$ (formula \eqref{pa}) and $\gamma^*(j,i)$ is the optimal scaling factor obtained from the $PA(\alpha_j, \alpha_i)$ (formula \eqref{pa}). 
Whenever $i=j$ we put
$$
pa(i,i)=0, \ \ \ \ dc(i,i)=0,  \ \ \ \ \gamma(i,i)=0.
$$
The numbers calculated by the formulas \eqref{padist}, \eqref{dcdist} and \eqref{scaldist}  can have very different ranges, hence, to ensure the comparability, we use the following normalization formulas for $dc$ and $\gamma$
$$
ndc(j)=\frac{1}{max_{i=1,...,n}{dc(i,j)}}, \ \ \ \ \text{for}\ \ \ j=1,...,n.
$$
$$
n\gamma(j)=\frac{1}{max_{i=1,...,n}{\gamma(i,j)}}, \ \ \ \ \text{for}\ \ \ j=1,...,n.
$$
When the normalization  is not needed we put $ndc(j)=ndc=1$, for all 
$j=1,...,n$ or $n\gamma(j)=n\gamma=1$, for all $j=1,...,n$.

\subsection{Similarity matrix}
\label{section_similarity_matrix}

The similarity matrix is created by calculating the similarity measure between any two contours $i$ and $j$. Let $\mu, \lambda, \omega\in \mathbb{R}$ be given numbers (weights), we calculate the similarity measure (SM) as follows:
\begin{equation}
\label{sm}
    SM_{ij}(\mu,\lambda,\omega,ndc, n\gamma):=
    \end{equation}
    $$
    \mu \cdot pa(i,j) +\lambda \cdot ndc(j) \cdot dc(i,j)+\omega \cdot n\gamma(j)\cdot \gamma(i,j).
$$

By calculating the number $SM_{ij}$ for every pair of contours $i$ and $j$, $i=1,...,n$, $j=1,...,n$  from the data set  we get the similarity matrix
\begin{equation}
\label{smmatrix}
SM(\mu,\lambda,\omega,ndc, n\gamma),
\end{equation}
which is symmetric and has zeros on the main diagonal. 

This general similarity measure allows us to preform various experiments. By changing values of numbers $\mu,\lambda,\omega,ndc, n\gamma$ we get the following similarity matrices:
\begin{enumerate}
\item  Procrustes similarity matrix 
\begin{equation}
\label{psm}
PSM=SM(1,0,0,1,1).
\end{equation}
\item Direct composition matrix
\begin{equation}
\label{dcm}
DCM=SM(0,1,0,1,1).
\end{equation}
\item Scale component matrix
\begin{equation}
\label{scm}
SCM=SM(0,0,1,1,1).
\end{equation}

\item Weighted Procrustes and scale component matrix
\begin{equation}
\label{wpsm}
 WPSM=SM(\mu,0,\omega,1,1).
\end{equation}
\item  Weighted direct  composition and scale component matrix where direct composition values are normalized
\begin{equation}
\label{wndc}
WNDCSM=SM(0,\lambda,\omega,ndc,1).
\end{equation}
\item Weighted direct composition and scale component matrix where both direct composition values and scale component values are normalized 
\begin{equation}
\label{WNDCNSM}
WNDCNSM=SM(0,\lambda,\omega,ndc,n\gamma).
\end{equation}
\end{enumerate}
For clustering  we apply the hierarchical clustering. We use three standard methods  to measure the distance between clusters
i.e single linkage,  average linkage and weighted average linkage (see Table \ref{tab_coph_org} and Table \ref{tab_coph_aug}).

\subsection{Proposed algorithm}
The general  algorithm  is presented below in Algorithm \ref{sh1}. By changing the values of weights $\mu, \lambda, \omega$ and $ndc$, $n\gamma$, we obtain different    similarity measures, as described above, and consequently, different clustering results. 

\begin{algorithm}
\label{sh1}
 	{\bf [1]}  Choose the values of weights $\mu, \lambda, \omega$ and $ndc$, $n\gamma$.
 	\smallskip
 	
 		{\bf [2]} Prepare the input data (vectors) as described in  Section \ref{section_classofobjects}. 
 		
 		\smallskip
 		{\bf [3]} Calculate the similarity between any two vectors $i$ and $j$  by the formula \eqref{sm} and generate the similarity matrix by the formula \eqref{smmatrix}.
 			
 		\smallskip
 		{\bf [4]}  Perform clustering  on the basis
 		of the similarity matrix \eqref{smmatrix} generated in step [3] by the standard hierarchical  algorithms  (Matlab toolbox) and generate the dendrogram.
 		
 \end{algorithm}

\section{Application-Archaeological objects.}
\label{section_application}

In the present section, we describe the archaeological objects to which we apply our algorithm. These objects are archaeological pottery fragments. When produced on the pottery wheel, vessels are rotationally symmetric objects of revolution. We limit our attention to this kind of vessels. Consequently, a cross-section and the location of the revolution axis are enough to convey the shape of the whole vessel. 

 In archaeological practice, the vessels or vessel fragments are described by their cross-sections. 
These cross-sections are hand-made technical drawings- they are made according to a set of rules and use conventions and a visual language to maintain standards and convey as much information about the vessel as possible, without the need to supplement it with text.

Classification of pottery obtained through excavation is a form of organizing the material to conclude the investigated sets and deal with the spatio-temporal diversity of the pottery. Such classifications are traditionally performed manually and depend heavily on the expert's knowledge and experience and as a result, are prone to being biased.

 In our application, the curves, as defined in Section \ref{section_classofobjects}, are contours of cross-sections of vessels uncovered in the course of archaeological work. 
From the technical drawings, the axis of revolution of each vessel can easily be deciphered. As mentioned in  Section \ref{section_classofobjects} to perform the clustering (generate clusters) we standardise the location of the contours to the first quadrant of $\mathbb{R}^{2}$ and standardise the position of contours by considering their upside-down versions. More precisely, in the present study we consider contours (curves) for which we know the coordinate $x_{r}$, see Figure \ref{fig:1} and for which the uppermost point (points) lies on the revolution axis OY to satisfy assumptions (a)-(c) of Section \ref{section_classofobjects}. This means that we classify vessel fragments for which the full cross-section is known. This does not mean that the full vessel is known. There exist archaeological techniques which allow reconstructing the exact full cross-section of a vessel, even in the case where the vessel is considerably damaged. 

During the process of clustering, the vessels can be divided into groups based on many features. With the help of our clustering algorithm, we divide given archaeological pottery sets with respect to shape and size. We limit our attention to the clustering of those pottery fragments for which the complete cross-section can be extracted.

\section{Experiment}
\label{section_experiment}

\begin{figure}
  \centering
  \includegraphics[width=6cm]{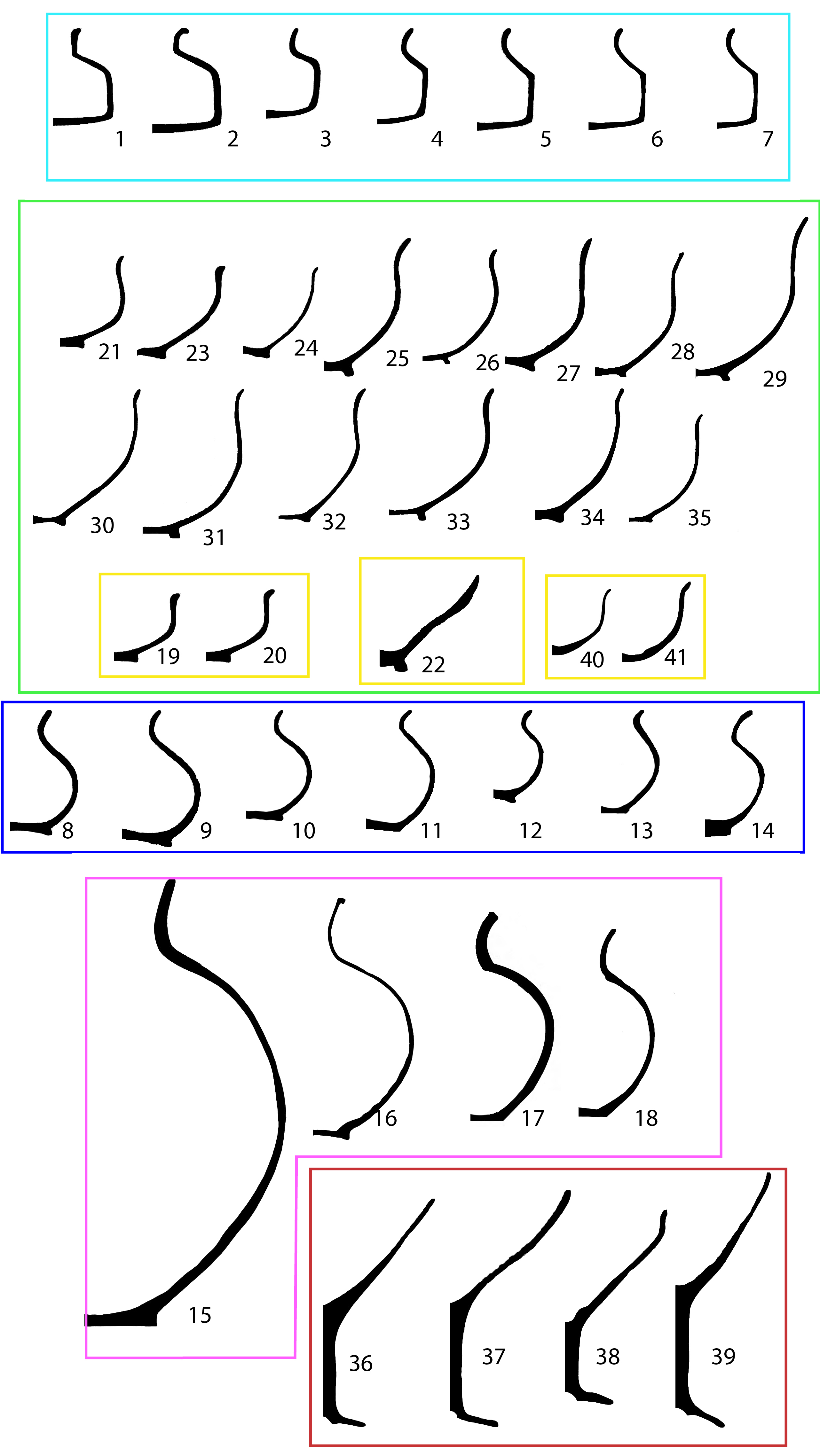}
  \caption{Set 2.}  \label{fig_set2}
\end{figure}

To test the algorithm  \eqref{sh1} we have designed and conducted a series of experiments related to the clustering of archaeological pottery. 
For that purpose, we have chosen 6 sets of data, each with 36 to 51 elements. Set 2 is presented in Fig.\ref{fig_set2}. 
The profiles used for the experiments come from the publication by P. Mountjoy \cite{mountjoy}  where the real-life archaeological pottery fragments are classified by experts. 

 In Table \ref{Tab_sets} we summarize the features of the sets 1-6. The estimated number of clusters is given as a range, as it depends on the method of counting the clusters (clusters vs. sub-clusters), and on the expert's decision. The table also provides a general description of the properties for each set, listing how the elements vary in size, and how much variation in shape there is within each cluster. 
\begin{table}[!h]
	\caption{Description of data-sets used in the experiment}  \label{Tab_sets}
\begin{tabular}{|p{0.9cm}|p{1.4 cm}|p{1.5cm}|p{3.7cm}|}
	\hline
	 Set Nr	&  Nr of elements &Estimated nr of clusters  &  Characteristics  \\
		\hline
Set 1	& 41  & 7-9 &  2 size groups, little variation in shape in clusters \\
    \hline
Set 2	& 41 & 6-10 &  2 size groups, at least 2 clusters with significant variation in shape in clusters \\
	\hline
Set 3	& 45 & 6-8 & 3 size groups, some variation in shape in clusters \\
    \hline
Set 4	& 36 & 5-7 &  all similar size, 4 basic shape groups, significant variation in shape in the clusters \\
	\hline
Set 5	&  51& 8-10& 5 size groups, significant differences in size between groups, some variation in shape  in the clusters  \\
    \hline
Set 6	& 40 & 5-8 &  similar size, 2 size groups, little variation in shape in clusters. \\
	\hline

		\end{tabular}
		\end{table}
	
The results of clustering obtained by algorithm \ref{sh1}  are visualized in the form of dendrograms.
The dendrogram's visual nature makes it inefficient when it comes to comparing  results for large data sets. 

In order to evaluate the quality of clustering, we use the cophenetic correlation coefficient to calculate a score for each clustering method and set, see Table \ref{tab_coph_org} and Table \ref{tab_coph_aug}.
The cophenetic correlation coefficient is a measure of how closely the dendrogram represents the pairwise dissimilarities between the objects. It was originally introduced by \cite{Sokal1962} as a method for comparing dendrograms resulting from numerical taxonomic research. 
The cophenetic correlation coefficient $c$ is defined as follows: 
\begin{equation}
\label{cophenet}
c := \frac{\sum_{i<j} (Y_{ij}- y) (Z_{ij}- z)}{\sqrt{\sum_{i<j} (Y_{ij}- y)^{2}\sum_{i<j}(Z_{ij}- z)^{2}}},
\end{equation}
where $Y_{ij}$ is the distance between objects $i$ and $j$ in the similarity matrix $Y$, and $Z_{ij}$ is the cophenetic distance, that is the distance between two observations $i$ and $j$ represented in a dendrogram by the height of the link at which those two observations are first joined. $y$ and $z$ are the average of Y and Z, respectively. The closer the value is to 1 the higher the quality of the solution. 
For a detailed discussion of the cophenetic correlation coefficient, please consult \cite{Farris1969}. 

\subsection{Augmentation}
\label{aug}

 The sets of archaeological pottery excavated by archaeologists are usually large and very diverse in terms of shape. Traditionally, the clustering (generating of typologies) is done manually by experts. Due to this,  in practice, only a small part of the archaeological material is clustered and published, as it is in    \cite{mountjoy}.

\begin{figure}
  \centering
  \includegraphics[width=7cm]{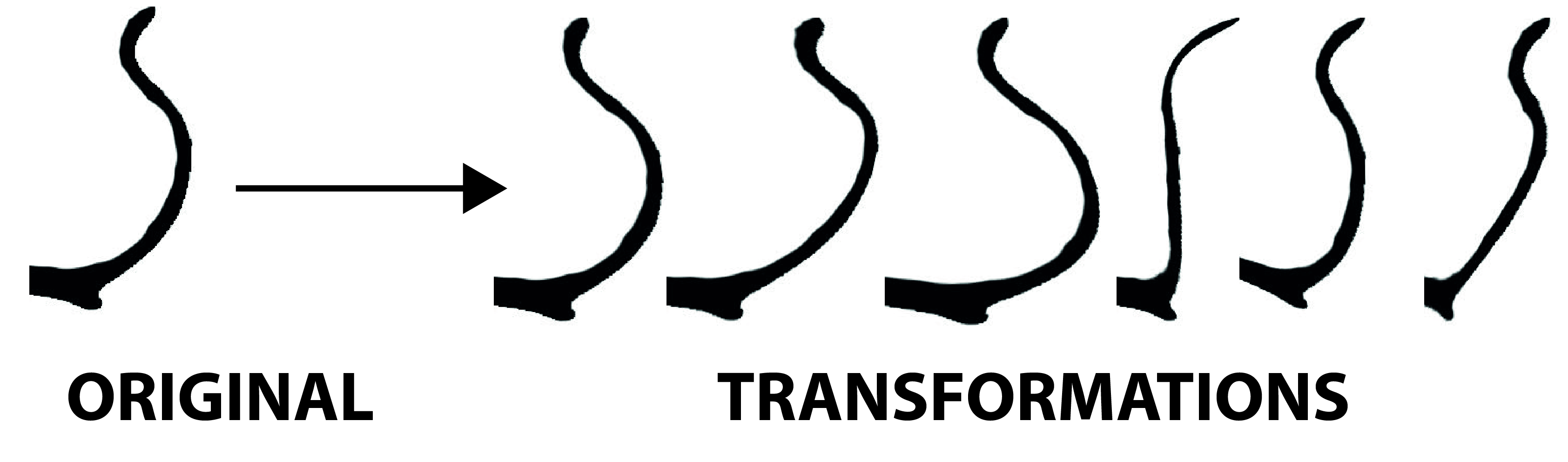}
  \caption{Illustration of the transformations applied in the augmentation process.}  \label{fig_augmentacja}
\end{figure}

 Hence, the real-life data sets which are available consist of 30-50 elements. This size does not provide a large enough sample to evaluate the performance of the algorithm. This was remedied by an automated augmentation of the data set. To ensure the comparability of the results we used the same data-sets (see Table \ref{Tab_sets}). Each element of the set has been subjected to 6 different transformations, resulting in sets with 252 to 357 elements (including the original objects). For these transformations, we have chosen warp type transformation, as they can produce slight changes, that are characteristic of our investigated elements. The warp transformations were grid-based and used a B\'{e}zier curve based grid to perform the transformations. Each of the 6 transformations targeted one part of the image (very generally described as top, middle, and bottom), and either bevelled the image towards the left or right edge of the image in the specific section. The transformations were chosen to be radical enough to modify the original image, to generate new groups, without ‘damaging' the image to the degree that it does not fulfil the original requirements (Fig.\ref{fig_augmentacja}) (see Section \ref{section_classofobjects}). Same as for the original data sets, the proposed methods have been tested on all 6 augmented sets, including different methods of the linkage algorithm. The results are included in Table \ref{tab_coph_aug}. 

\section{Results and Discussion}
\label{results_discussion}

In this section, we discuss the results of the experiments. 

The evaluation of the results is based on the dendrograms obtained for each set. The cophenetic correlation coefficient (Tables \ref{tab_coph_org} and \ref{tab_coph_aug}) is used as an additional measure, evaluating the quality of the dendrogram. 

The crucial point in this kind of clustering problem is the definition of similarity measure which should reflect the similarity as it is perceived by the expert. The following properties of the proposed similarity measures can be observed based on the experiment:

\begin{enumerate}

\item First type of mismatch occurs when two contours are very similar or even identical on a long portion of the curve and differ significantly on a relatively short portion of the curve. 
In such a case, the high similarity, i.e. the small distance between curves on a given portion {\it{overpowers}} the similarity measure calculated for the two objects, and fails to take into account the difference between the objects.  This can lead to a false-positive result, with two contours perceived as "different", and belonging to different clusters, being grouped.

\item The second type of mismatch in the dendrogram can occur when the weights are improperly selected. This means that one of the aspects of the clustering, the shape or scale, takes precedence over the other and leads to two contours being paired or separated as e.g. their scale is very similar, despite not exhibiting similarity in shape. 

\item The third type of mismatch stems from the difference between the perceived similarity (experts evaluation) and the calculated similarity measure. The in-depth discussion of this is outside of the scope of this paper, however, it will be demonstrated below in the discussion of the results obtained for the original data-sets (Section \ref{results for orginal data}). 

\end{enumerate}

Those types of mismatches do not point to a fault in the method. They are an inherent property of automated shape detection, especially Type 1. Type 2 can be remedied by the appropriate selection of weights by the expert. These types of mismatches highlight the discrepancies between manual {\it{handmade}} classification, where consistency of decision making is hard to maintain all through the process and personal biases come into play. Whereas with an automated process the same criteria are applied throughout. The same applies to Type 3.

\subsection{Results obtained for the original data-sets}
\label{results for orginal data}
 The analysis of the dendrograms has yielded the best results for the combination of the Mix method and the scale component. The exact choice of weights depends on the nature of the set. The WNDCSM similarity measure turned out to be the best choice for sets 1, 3, 4, 5, and 6. This is due to the differences in shape coinciding with the differences in scale. That is, there are few or no contour fragments that overlap in similar shape and similar size, although belonging to different clusters. For set 2 the best combination is weight 3/4 for Procrustes and 1/4 for the Scale Component. For this set, the weight of the Scale Component had to be reduced in favour of the shape component PA of the measure. This is due to several contours being {\it{close}}, both in terms of shape and size, and the need to put more stress on the shape component, as this is our primary goal.
 
The results of an expert-based evaluation of the obtained dendrograms are summarized in Table \ref{tab_results}. The table presents the total number and percentage of the elements that are, in the expert's opinion, correctly classified. 
As the tool is meant to be of assistance to experts, we choose an expert-based evaluation for our results. An automated or semi-automated assessment of results based on the dendrograms, and the choice of a correct similarity threshold is itself a matter of ongoing research (e.g. see \cite{MLCut}, \cite{AMCS2}. 

\begin{figure*}
  \centering
  \includegraphics[width=14cm,height=4.85cm]{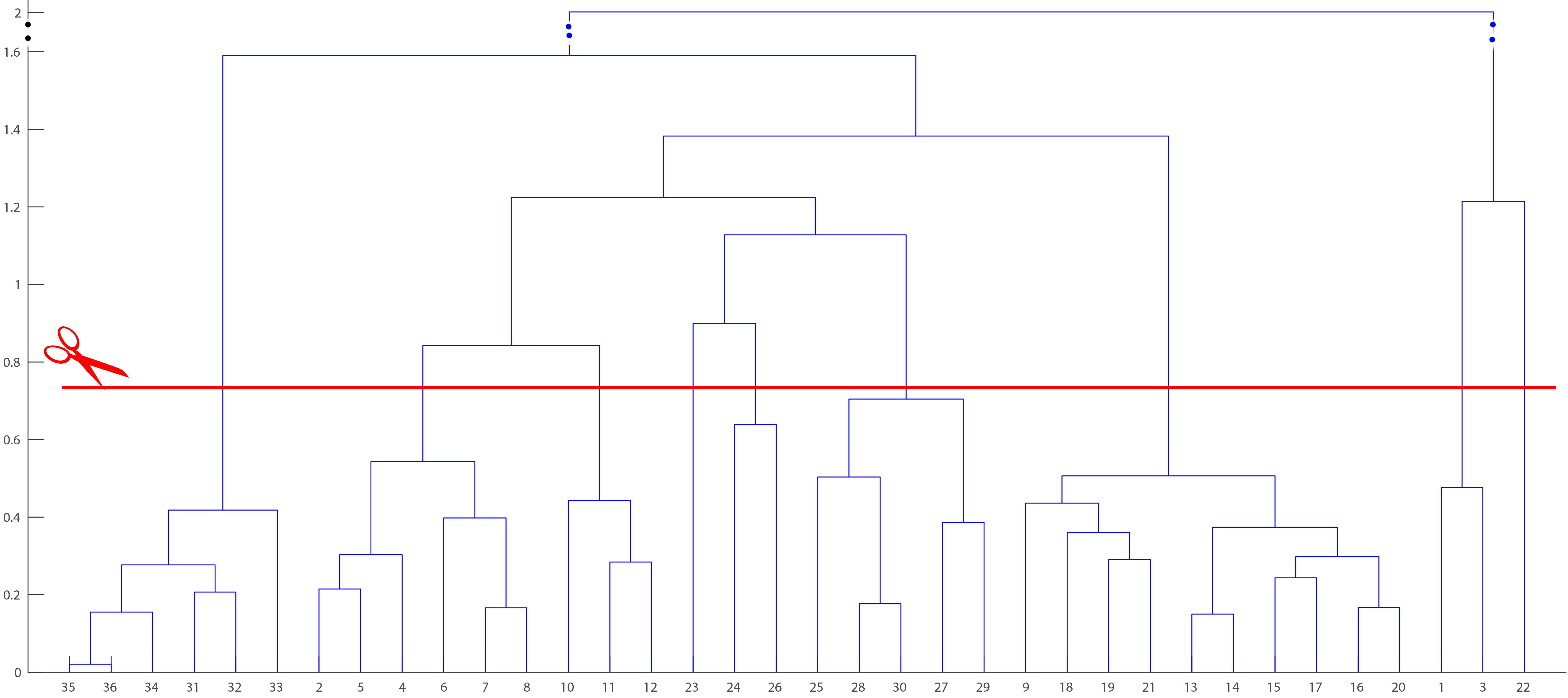}
  \caption{A single-level cut through the dendrogram obtained for the Set 2 according to formula \ref{wndc}. CCF=0,98, correctly classified objects= 83,34\%.}  \label{Set_4_single cut}
\end{figure*}

\begin{figure*}
  \centering
  \includegraphics[width=14cm,height=4.85cm]{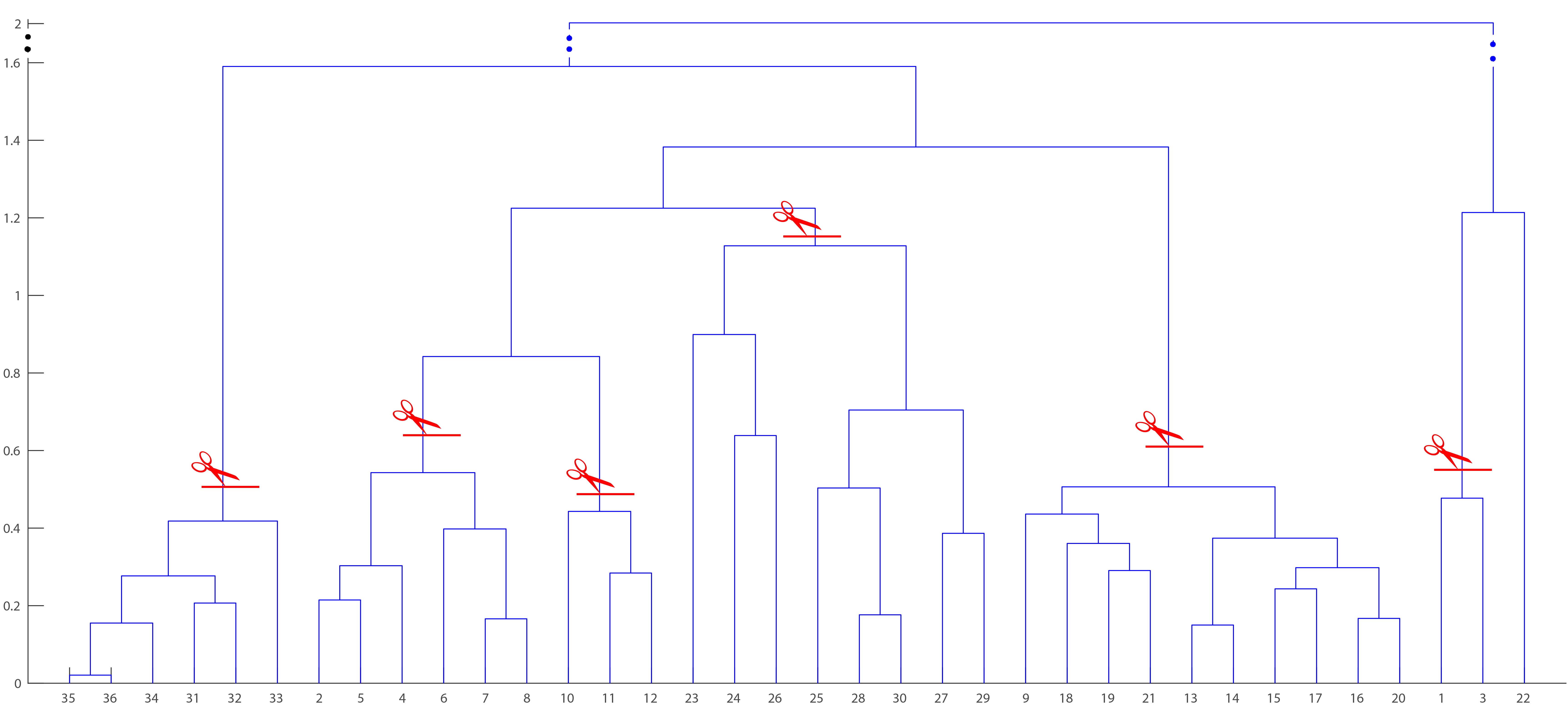}
  \caption{A Multi-level cut through the dendrogram obtained for the Set 4 according to formula \ref{wndc}. CCF=0,89, correctly classified objects= 91,67\%.}  \label{Set_4_multi cut}
\end{figure*}

A visualization of the expert's assessment is presented in Figures \ref{Set_1_single cut}, \ref{Set_1_multi cut}, \ref{Set_4_single cut}, and \ref{Set_4_multi cut}. For dendrograms that have a smaller similarity measure values between objects and groups, the difference between the single cut and the multi-level cut is not significant. For Set 2 (Figures \ref{Set_1_single cut}, \ref{Set_1_multi cut}), for the single-level cut we obtain the accuracy of 80\% (compared to 92,68 \% with the multi-level cut-\ref{tab_results}). The dendrogram has the CFF value of 0,98, which means the dendrogram's structure is very good. For Set 4 (Figures \ref{Set_4_single cut}, \ref{Set_4_multi cut}), CFF value is significantly lower, with 0,89. This dendrogram has higher values of similarity measures between objects. Using the multi-level cut method we obtain the result of 91,67 \%. The single-level cut method results in only 83,34 \% of accuracy.  The amount of correctly classified elements and the resulting percentages, as presented below, have been calculated based on the expert's judgement. In calculations of the results of Table 2, the cuts of the dendrograms have been made on different levels (multi-level cuts), i.e. different thresholds (cut levels) selected for splitting the dendrogram (i.e. obtaining clusters) have been chosen by the expert and are based on the direct data inspection.  

The expert-based evaluation is supplemented by  the  cophenetic correlation coefficient (CCF) (Table \ref{tab_coph_org}) calculated for each set and each method in the experiments performed. A discussion of the obtained results, summarized in Table \ref{tab_results} and Table \ref{tab_coph_org} is presented below, in Section \ref{section_disscusionofresults}.  

\subsection{Discussion of the results}
\label{section_disscusionofresults}
\begin{figure*}
  \centering
  \includegraphics[width=14 cm, height=4.85cm]{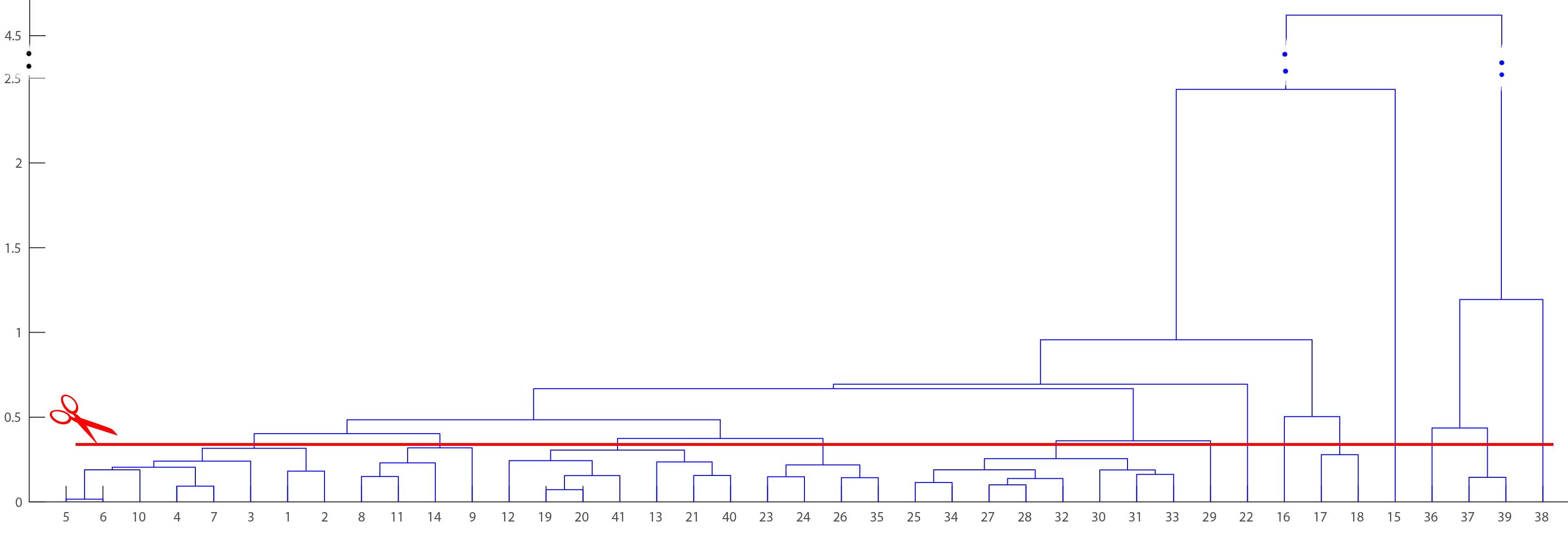}
  \caption{A single-level cut through the dendrogram obtained for the Set 2 according to formula \ref{WNDCNSM}. CCF=0,98,  correctly classified objects= 80\%.}  \label{Set_1_single cut}
\end{figure*}

\begin{figure*}
  \centering
  \includegraphics[width=14 cm, height=4.85cm]{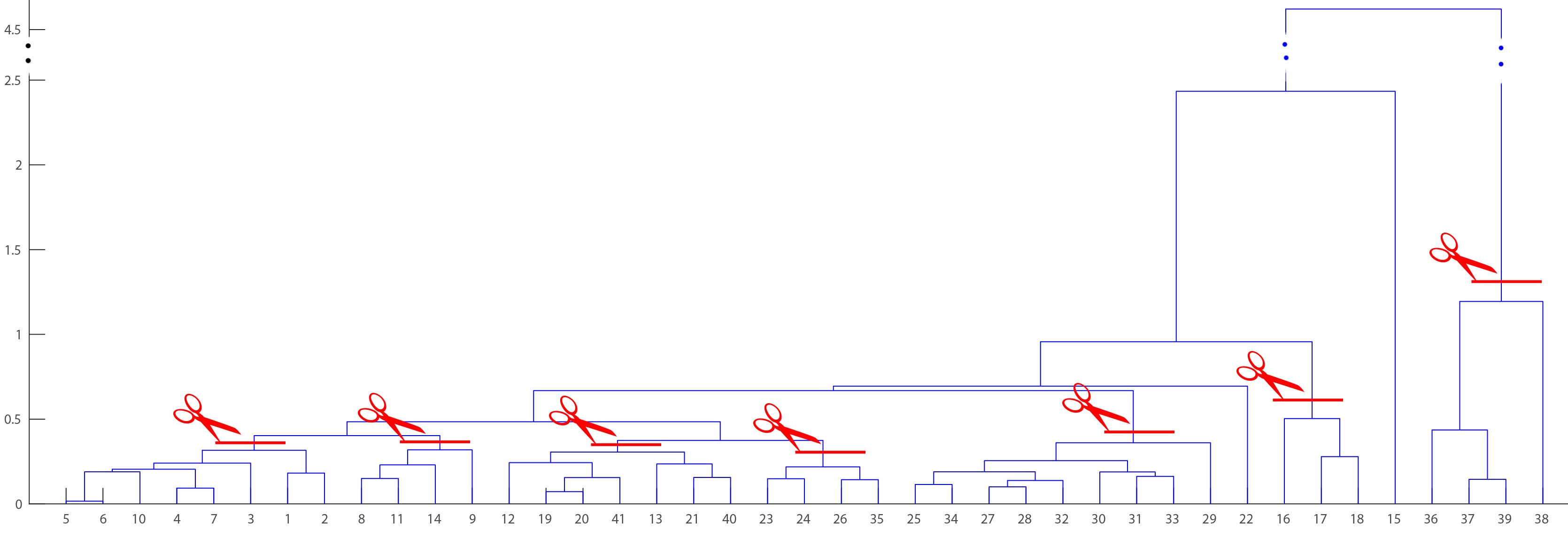}
  \caption{A Multi-level cut through the dendrogram obtained for the Set 2 according to formula \ref{WNDCNSM}. CCF=0,98, correctly classified objects= 92,68\%.}  \label{Set_1_multi cut}
\end{figure*}

One of the results, for the WNDCSM  similarity measure, obtained through the experiments is presented in Figures  \ref{Set_1_single cut} and  \ref{Set_1_multi cut}.

Based on Table \ref{tab_results} we notice that the best results were obtained for the method WNDCNSM ($SM(0,\frac{3}{4},\frac{1}{4},ndc,n\gamma$) for sets 1, 2, 5, and 6. This can be attributed to the character of the sets, as summarized in Table \ref{Tab_sets}. Sets 3 and 4 show significant variation in shape within the clusters, which might result in some under-performance of the method. However, the results are still fairly high. Combining together the results in Table \ref{tab_results} and Table \ref{tab_coph_org} we see that all proposed methods perform fairly well, as the performance is on average 85\% (Table \ref{tab_results}) for the 3 proposed methods, and all values for CCF are at 0,75 and above (Table \ref{tab_coph_org}), with most of them being above 0,9. As stated before, the values of CCF do not evaluate the clustering results, but rather the structure of the dendrogram, and in this sense, the performance of the algorithm.

The results as described above prove the efficiency of the WNDCSM algorithm, and the introduction of the additional DWT distance component to the similarity measure. The results point to a double function of the WPSM similarity measure and all the subsequent combinations, in the sense of Procrustes analysis being responsible for a robust detection of the "global shape" while DTW detects subtle differences between shapes, and allows taking into account of the scaling factor. Furthermore, the Scaling component accounts for the difference in size, between the investigated objects. Such a composition allows for a fuller analysis of the given sets.

\subsection{Results obtained for the augmented data-sets}

The augmented data sets were clustered by the algorithm in Section \eqref{sh1}, the same as the original data. Due to the method of augmentation (Section \ref{aug}), each set increased in the number of shapes, however, no new shape groups were created, as the augmentation did not target size, solely the shape. Due to the size of the resulting dendrograms, we omit here the detailed discussion. However, the satisfactory performance of the algorithm \eqref{sh1} for the original sets, and the high values of the cophenetic correlation coefficient, gives an insight into the performance of the algorithm for the augmented sets. The augmentation process will be applied in future research to increase the size of data sets to employ machine learning algorithms.

\section{Conclusions}
\label{section_conclusions}
 We have presented the conception and performance of the algorithm (see Section \eqref{sh1}) for clustering of a class of 2D objects, with application in archaeology. The performed experiment shows that the proposed algorithm provides satisfactory results. Even if, the results obtained are promising, many problems related to the clustering of archaeological pottery remain open. 

Conceivable extensions of the research presented here encompass investigating the possible advantages of using alternative shape representations and similarity measures.  As a part of our further research, we plan to investigate the case where only a fragment of the section is preserved. Another problem arises when the investigated pottery vessel has a handle. As far as we know this problem has not yet been addressed in the literature. 

\begin{table*}[!t]
 \caption{Percentage of correctly classified elements as evaluated by the expert for Sets 1-6. The number of elements in each set, and the number of correctly classified elements, is given in brackets. All scores have been calculated using the average distance between clusters.}
\label{tab_results}
\renewcommand{\arraystretch}{1.4}
\begin{tabular}{|c|c|c|c|c|c|c|}
\hline
Method   & set 1 [41]& set 2 [41] & set 3 [45]& set 4 [36] & set 5 [51] & set 6 [40] \\
\hline

$SM(\frac{1}{2},0,\frac{1}{2},1,1)$
	 & 82.93 \%[34]& 78.05\%[32]& 75.56\%[34]&  75\%[27] & 69.23\%[36]& 87.5\%[35]\\

 	\hline
 $SM(\frac{3}{4},0,\frac{1}{4},1,1)$ 
 & 85.36\% [35]& 90.24\%[37] & {\bf 82.22\%[37]}&  86.11\%[31] &78.43\%[40]& 85\%[34]\\
	
	\hline
$SM(0,\frac{1}{2},\frac{1}{2},ndc,1)$  
	 &  {\bf90.24\%[37]}& 80.49\%[33] & 77.78\%[35] &  86.11\%[31]& 78.43\%[40]& 82.5\%[33]\\
	
		\hline
	$SM(0,\frac{3}{4},\frac{1}{4},ndc,1)$  
	& 87.8\%[36]& 90.24\%[37]&  77.78\%[35]&  {\bf 91.67\%[33]}& 78.43\%[40]&  87.5\%[35]\\

		\hline
$SM(0,\frac{1}{2},\frac{1}{2},ndc,n\gamma)$  
	 &  87.8\%[36]& 75.61\%[31]& 75.56\%[34]& 88.89\%[32]& 78.43\%[40]& {\bf 92.5\%[37]}\\

		\hline
	$SM(0,\frac{3}{4},\frac{1}{4},ndc,n\gamma)$   
	 & {\bf 90.24\%[37]}& {\bf 92.68\%[38]}& 80\%[36]& 86.11\%[31]& {\bf 82.35\%[42]}& {\bf 92.5\%[37]}\\

		\hline
\end{tabular}
\end{table*}
\renewcommand{\arraystretch}{1}
\begin{table*}[!t]
 \caption{Cophenetic correlation coefficient calculated for 6 investigated data-sets.}
\label{tab_coph_org}
\begin{tabular}{|c|p{1.5cm}|c|c|c|c|c|c|p{1.4cm}|}
	\hline
	
Method  &  & set 1 & set 2 & set 3 & set 4 & set 5 & set 6 & Arithmetic average\\
\hline
$SM(1,0,0,1,1)$	&  & & & &  & & &\\
\hline
	& single & 0.8564 &   0.9952 & 0.9396 & 0.6340 & 0.5906
 & 0.7495 & 0.7942\\
	\hline
	& average  & {\bf  0.9958}  & 0.9221 & 0.9194  &  0.7930 & 0.8141 & 0.8546 & 0.8832\\
	\hline
	& weighted  & 0.9189 & {\bf 0.9956}  &  0.9047  &  0.7838 & 0.7692 &  0.8456 & 0.8696 \\
	\hline 
	$SM(0,1,0,1,1)$	&  &  &  & & & & & \\
	\hline
		& single & 0.8549 &  0.9872 &  0.9492 &  0.8325 & 0.8165& 0.7894&0.8716 \\
	\hline
	& average & 0.9082 &  0.9892 & 0.9735 &   0.8446 & 0.8966 & 0.8328&0.9075\\
	\hline
	& weighted & 0.9038  & 0.9886 &  0.9725 &  0.8426 &0.8598 &  0.8172& 0.8974\\
	\hline
$SM(0,0,1,1,1)$	&  & & & & & & &\\
\hline
		& single  & 0.9760&  0.8939 &0.9964 &  0.8973 & 0.9155 & 0.9056&0.9308\\
	\hline
	& average &  0.9792  & 0.9379 & {\bf 0.9974} &   0.9423 &  {\bf 0.9293} & 0.9170&0.9505
\\
	\hline
	& weighted  &  0.9774 & 0.8730 & 0.9961 &  0.9404 & 0.9284 & 0.8993 &0.9358\\
		\hline
$SM(\frac{1}{2},0,\frac{1}{2},1,1)$ &  & & & & & && \\
	\hline
		& single & 0.9645 &  0.9747&0.9874 &
		0.8816 & 0.9100 & 0.9467 &0.9441 \\
	\hline
	& average & 0.9705 & 0.9815 &0.9881 & {\bf 0.9450} &0.9256& {\bf 0.9542}&{ \bf 0.9608}\\
	\hline
	& weighted & 0.9681 & 0.9803 & 0.9880 &  0.9414 & 0.9248&  0.9532&0.9593
 \\
 	\hline
 $SM(\frac{3}{4},0,\frac{1}{4},1,1)$ & &  & & & & && \\
	\hline
		& single & 0.8963 & 0.9917 & 0.9446 & 0.8986 & 0.8915 &  0.9150&0.9229 \\
	\hline
	& average & 0.9287 &  0.9933  &  0.9450 &  0.9298 &0.9068& 0.9292 &0.9388\\
	\hline
	& weighted & 0.9268  &  0.9933 & 0.9428 &  0.9386 & 0.9061& 0.9262&0.9390\\
	\hline
$SM(0,\frac{1}{2},\frac{1}{2},ndc,1)$ &  & & &  & & & &\\
	\hline
		& single& 0.9441 &  0.9670 & 0.9704 &   0.8023 & 0.8034& 0.9110&0.8997\\
	\hline
	& average &  0.9611  & 0.9583 &0.9717  &  0.8862 &0.8991 & 0.9199&0.9327 \\
	\hline
	& weighted & 0.9588 &  0.9724 & 0.9708  &  0.8763 &0.8954 & 0.9188& 0.9321\\
		\hline
	$SM(0,\frac{3}{4},\frac{1}{4},ndc,1)$ &  & & & & & && \\
	\hline
		& single &  0.9042 &   0.9806 &0.9353 &  0.8189 &0.7574 & 0.8456&0.8737 \\
	\hline
	& average& 0.9309  & 0.9845 &  0.9368 &  0.8279 & 0.8739&  0.8641 & 0.9030\\
	\hline
	& weighted & 0.9285 &  0.9828 & 0.9360 &  0.8463 &   0.8707&  0.8561&0.9034\\
		\hline
$SM(0,\frac{1}{2},\frac{1}{2},ndc,n\gamma)$ &  & & &  &  & &&\\
	\hline
		& single & 0.9513  &   0.9589 &0.9726  &   0.8147 &  0.8543 & 0.9248&0.9128\\
	\hline
	& average &  0.9650  & 0.9583 & 0.9738 & 0.9030 & 0.9125& 0.9372&0.9416 \\
	\hline
	& weighted & 0.9626 &  0.9535& 0.9734  &  0.9029 & 0.9047& 0.9355&0.9388\\
		\hline
	$SM(0,\frac{3}{4},\frac{1}{4},ndc,n\gamma)$ &  &  & & & & & &\\
	\hline
		& single & 0.9116  &   0.9784 &   0.9372 &  0.8177 &0.7518 & 0.8610 &0.8763\\
	\hline
	& average & 0.9343 &  0.9836 &  0.9387&  0.8509 & 0.8855 & 0.8762&0.9115 \\
	\hline
	& weighted & 0.9361 &  0.9817 &   0.9378
&  0.8365 &0.8828 & 0.8746&0.9083 \\
		\hline
\end{tabular}
\end{table*}
\begin{table*}[!t]
\caption{Cophenetic correlation coefficient results for the augmented data-sets.}
\label{tab_coph_aug}
\begin{tabular}{|c|p{1.5cm}|c|c|c|c|c|c|p{1.4cm}|}
	\hline
	
Method &  & set 1a & set 2a & set 3a & set 4a & set 5a & set 6a & Arithmetic average\\
\hline
$SM(1,0,0,1,1)$	&  & & & &  & & &\\
\hline
	& single & 0.6780 &  0.9788&  0.7984 & 0.5813 & 0.6211
 &  0.1760 &0.6389 \\
	\hline
	& average  & 0.8777  & 0.9836 & 0.8968  & 0.7844 & 0.8088 & 0.7933&0.8574 \\
	\hline
	& weighted  & 0.8530 & 0.9808 &  0.8583  &  0.8042 & 0.7633 &  0.7536&0.8355 \\
	\hline 
	$SM(0,1,0,1,1)$	&  &  &  & & & & & \\
	\hline
		& single & 0.8520 &  0.9736
 &   0.9454 &  0.7676 & 0.8182& 0.6921 & 0.8415\\
	\hline
	& average & 0.8972 &  0.9771 & 0.9584 &   0.7852 & 0.8599 & 0.7548&0.8721\\
	\hline
	& weighted & 0.9104  & 0.9617 &  0.9750 &  0.6916 &0.8505 &  0.7464&0.8559\\
	\hline
$SM(0,0,1,1,1)$	&  & & & & & & &\\
\hline
		& single  & 0.9592&  0.8019 &0.7192 &   0.6006 & 0.8912 & 0.8213&0.7989\\
	\hline
	& average & {\bf 0.9731}  & 0.8770 &  0.9613&  0.8664 & 0.9131 &{\bf  0.8602}& {\bf 0.9085}
\\
	\hline
	& weighted  &  0.9687 &  0.8517 & 0.9566 & {\bf 0.8742} & 0.7725& 0.7979&0.8703\\
		\hline
$SM(\frac{1}{2},0,\frac{1}{2},1,1)$ &  & & & & & & &\\
	\hline
		& single & 0.9439 & 0.9619&{ \bf 0.9874} &
	 0.7321 & 0.8515 & 0.6190& 0.8493  \\
	\hline
	& average & 0.9573 & 0.9713 &0.9091 &  0.8688& {\bf 0.9009}&  0.7758&0.8972 \\
	\hline
	& weighted & 0.9485 &  0.9513 & 0.8162 &  0.6914 & 0.8706& 0.7385&0.8361
 \\
 	\hline
  $SM(\frac{3}{4},0,\frac{1}{4},1,1)$  & &  & & & & & &\\
	\hline
		& single & 0.8457 & 0.9803 &0.7350 & 0.7211 &  0.8139 &   0.3865 &0.7471\\
	\hline
	& average & 0.8850 & {\bf 0.9825} &  0.8749&  0.8276 &0.8581& 0.7608 & 0.8648\\
	\hline
	& weighted & 0.7902  & 0.9804 & 0.8155 &  0.8077 & 0.8472& 0.7680&0.8348\\
	\hline
$SM(0,\frac{1}{2},\frac{1}{2},ndc,1)$&  & & &  & & & &\\
	\hline
		& single&  0.9428 &  0.9539 & 0.9046 &    0.7411 & 0.7916&  0.7523&0.8477\\
	\hline
	& average & 0.9537   &  0.9550 &0.9737  &  0.7923 &0.8566 & 0.8314 & 0.8938\\
	\hline
	& weighted & 0.9571 &  0.9351 & 0.9685
  &  0.7862 &0.8402 & 0.7957 &0.8805\\
		\hline
		$SM(0,\frac{3}{4},\frac{1}{4},ndc,1)$ &  & & & & & & &\\
	\hline
		& single &   0.9082 &      0.9701 &0.9562 &  0.7725 & 0.7410& 0.7258&0.8456 \\
	\hline
	& average& 0.9396  & 0.9720 & 0.9735 &  0.8261 & 0.8388&  0.7858&0.8893\\
	\hline
	& weighted & 0.9386 &  0.9660 & 0.9720 &  0.7718
&   0.8590&  0.7646& 0.8787\\
		\hline
$SM(0,\frac{1}{2},\frac{1}{2},ndc,n\gamma)$ &  & & &  &  & &&\\
	\hline
		& single & 0.9474  &   0.9459 &0.9008  &   0.7191 &  0.8154 & 0.7832&0.8520\\
	\hline
	& average &  0.9574  & 0.9556 & 0.9739 & 0.8208 & 0.8867 & 0.8519&0.9077\\
	\hline
	& weighted & 0.9536 &  0.9338& 0.9651  &  0.7628 & 0.8519& 0.7893&0.8761\\
		\hline
		$SM(0,\frac{3}{4},\frac{1}{4},ndc,n\gamma)$ &  &  & & & & & &\\
	\hline
		& single & 0.9104 &  0.9686 &    0.9552 &  0.7625 &0.7558 & 0.7232&0.8460 \\
	\hline
	& average &0.9431 &  0.9711 & 0.9737&  0.7773 & 0.8550 & 0.8012 &0.8869\\
	\hline
	& weighted & 0.9279  &  0.9631
 &    0.9721
&  0.7560 &0.8670 & 0.7957&0.8803 \\
		\hline
\end{tabular}
\end{table*}

\bibliographystyle{plain}
\bibliography{representation}

\end{document}